\documentclass[10pt,twocolumn,letterpaper]{article}

\usepackage[pagenumbers]{cvpr} 

\definecolor{cvprblue}{rgb}{0.21,0.49,0.74}
\usepackage[pagebackref,breaklinks,colorlinks,allcolors=cvprblue]{hyperref}

\def\paperID{****} 
\def\confName{CVPR}
\def\confYear{2026}

\title{Dynamic Stream Network for Combinatorial Explosion Problem in Deformable Medical Image Registration}

\author{Shaochen Bi\footnotemark[1]\\
Hong Kong University of Science and Technology
\\
Hong Kong, China\\
{\tt\small bisc0507@163.com}
\and
Yuting He\footnotemark[1]  \footnotemark[2]\\
Case Western Reserve University\\
Cleveland, OH 44106 USA\\
{\tt\small yuting.he4@case.edu}
\and
Weiming Wang\\
Hong Kong Metropolitan University\\
Hong Kong, China\\
{\tt\small wmwang@hkmu.edu.hk}
\and
Hao Chen\footnotemark[2]\\
Hong Kong University of Science and Technology\\
Hong Kong, China\\
{\tt\small jhc@ust.hk}
}

\begin{document}

\maketitle
\let\thefootnote\relax\footnotetext{
$^*$ These authors contributed equally \hspace{5pt}
$^\dagger$ Corresponding author \hspace{5pt}
}

\begin{abstract}
Combinatorial explosion problem caused by dual inputs presents a critical challenge in Deformable Medical Image Registration (DMIR). Since DMIR processes two images simultaneously as input, the combination relationships between features has grown exponentially, ultimately the model considers more interfering features during the feature modeling process. Introducing dynamics in the receptive fields and weights of the network enable the model to eliminate the interfering features combination and model the potential feature combination relationships. In this paper, we propose the Dynamic Stream Network (DySNet), which enables the receptive fields and weights to be dynamically adjusted. This ultimately enables the model to ignore interfering feature combinations and model the potential feature relationships. With two key innovations: 1) Adaptive Stream Basin (AdSB) module dynamically adjusts the shape of the receptive field, thereby enabling the model to focus on the feature relationships with greater correlation. 2) Dynamic Stream Attention (DySA) mechanism generates dynamic weights to search for more valuable feature relationships. Extensive experiments have shown that DySNet consistently outperforms the most advanced DMIR methods, highlighting its outstanding generalization ability. Our code will be released on the website: https://github.com/ShaochenBi/DySNet.
\end{abstract}    
\section{Introduction}
\label{sec:intro}

\begin{figure}[t]
  \centering
   \includegraphics[width=1\linewidth]{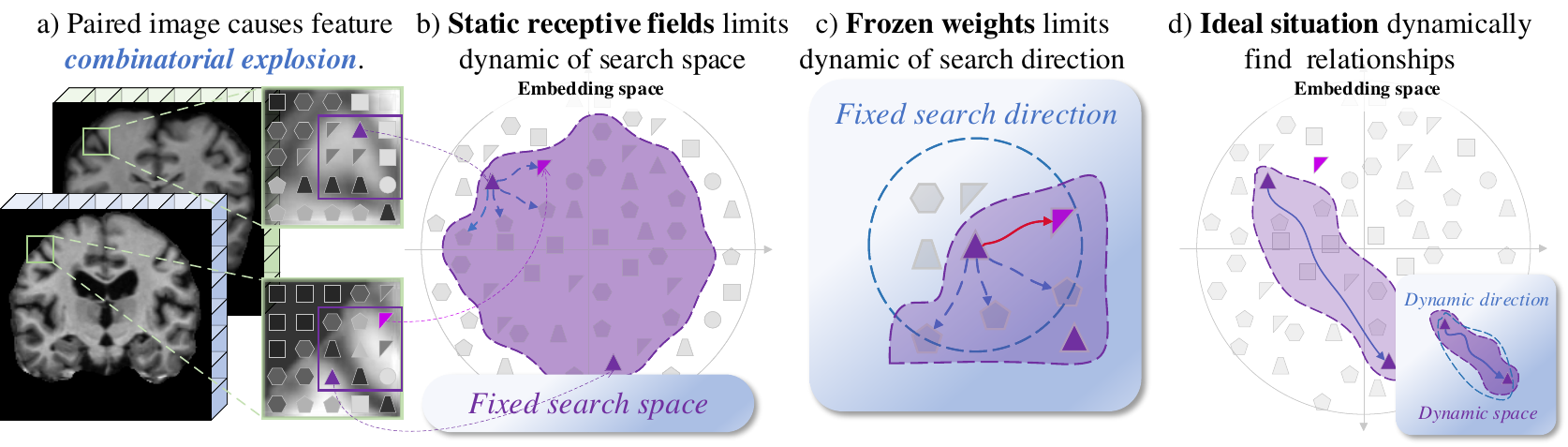}
   \caption{\textbf{Combinatorial explosion problem in DMIR}: (a) dual inputs cause a \textit{combinatorial explosion} of features. (b) static receptive fields contains a large number of interfering features. (c) static weights ignore potential accuracy features. (d) ideal situation where dynamic receptive fields and weights adaptively model the feature relationships.}
   \label{i1}
\end{figure}

Although feature modeling has long been a key research task~\cite{swin_voxelmorph,lku,transmorph,vit-v-net,modet,sacb,adamw,glu,xmorpher,voxelmorph} in deformable medical image registration (DMIR~\cite{DMIRsurvey}), it is still constrained by an open problem, \textit{combinatorial explosion}~\cite{deepNET,HU20181}. DMIR models the feature relationships between two images, which is different from traditional single image input tasks (such as segmentation~\cite{unet} and classification~\cite{he2016deep}). This results in an exponential increase in the combination of feature relationships, ultimately leading to the problem of feature relationship combination explosion. This problem leads to the model expanding the search space and direction to capture more relevant feature relationships (Fig.\ref{i1}-b). As the search space and direction expand, more irrelevant feature combinations relationships also be taken into consideration, causing the model to select suboptimal feature relationships~\cite{krebs2017robust, rohe2017svf, voxelmorph} (As the red line in Fig.\ref{i1}-c).

Existing feature modeling studies in DMIR~\cite{xmorpher,voxelmorph,swin_voxelmorph,lku,transmorph,vit-v-net,modet,sacb,adamw,glu} attempt to alleviate this problem. However, there are still two limitations. \textit{1) Static receptive fields limit the adaptability of relationship search space (Fig.\ref{i1}-b).} Existing registration networks~\cite{voxelmorph, lku, modet,swin_voxelmorph,xmorpher,vit-v-net} adopt static receptive fields, \eg convolution kernels~\cite{cnn}, patch partition~\cite{vit,swin_transformer} of Transformer. The static receptive fields fix the number and range of feature relationships considered for a single feature, thereby constraining the relationship search space to a fixed shape and structure. Therefore, the model must expand the search space with fixed shape to capture more accurate feature relationships~\cite{luo2016understanding,nonlocal}. This whole expansion of the relationship search space leads to an increase in interfering feature relationships in the relationship search space, ultimately affecting the accuracy of the feature modeling~\cite{he2020momentum}. \textit{2) Static weights will limit the flexibility of the relationship search direction (Fig.\ref{i1}-c)}. The existing registration networks use the static weights for testing after model training~\cite{voxelmorph, lku}. These static weights do not adjust their values according to the test data, thus the model perform inference under the static weights. This results in a fixed relationship search direction during feature modeling, preventing the capture of some potential feature relationships~\cite{yosinski2014transferable}. At the same time, if the relationship search space is expanded, the fixed search direction will overlook more potential feature relationships, ultimately affecting the performance of the model.

Although some preliminary registration networks \cite{sacb,xmorpher,transmorph,modet} have made efforts to incorporate dynamic mechanisms into either receptive fields or weights individually, none have successfully realized the dynamic adaptation of both components. This limitation restricts the model from flexibly conducting feature modeling in a unified manner.

\begin{figure}[t]
  \centering
   \includegraphics[width=1\linewidth]{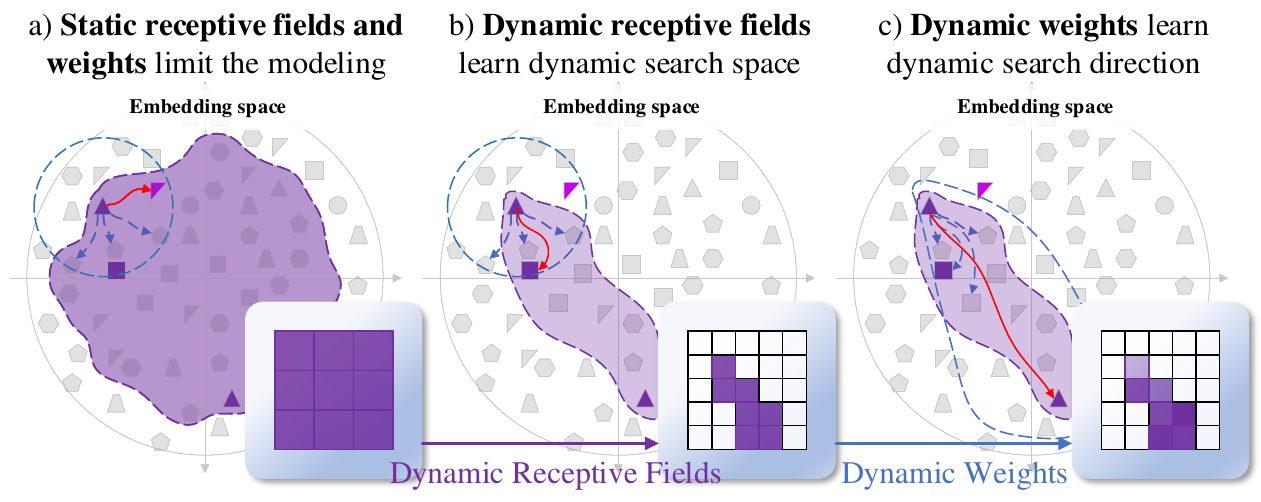}
   \caption{\textit{Motivation}: (a) shows that static receptive fields and static weights restrict learning ability by limiting the search space and direction. (b) illustrates dynamic receptive fields enable the adaptability of relationship search space. (c) demonstrates that dynamic weights allow for precise localization of feature relationship, enhancing matching accuracy by adjusting the search space and direction dynamically in the embedding space.}
   \label{i2}
\end{figure}

\textit{Motivation}: An intuitive solution is to make the receptive fields and weights dynamic according to the input images (Fig.~\ref{i1}-d). These two dynamics enable the network to dynamically adjust the receptive fields and weights, thereby modifying the relationship search space and direction during the feature modeling process. Ultimately, this allows the model to introduce as few interfering features as possible into the relationship search space while employing a more accurate relationship search direction for feature modeling. 1) \textit{Dynamic receptive fields}: This dynamic mechanism adjusts the shape of the window (such as deformable convolution \cite{dcn,dcn2} and deformable attention \cite{dat}) according to the input image, thereby adjusting the receptive fields of the model. The dynamic receptive fields enables the search space to adaptively model the feature relationships, allowing the model to reduce the number of interfering features in the search space (as shown in Fig.\ref{i2}-b, the static receptive field has filtered out a large number of interfering features through deformation) \cite{dat, sacb, dcn, dcn2}. 2) \textit{Dynamic weights}: This mechanism enables the model to modify the weight values (such as transformer \cite{attention}). This dynamic adjustment of the weight values (As shown in Fig.\ref{i2}-c, model searches for a specific direction by adjusting the weight values.) enables the model to change the range and depth of the search direction according to the similarity of the images, ultimately achieving adaptive search directions in the search space and capture potential feature relationships \cite{attention}.

In this paper, we propose a dynamic feature modeling network called ``Dynamic Stream Network" (DySNet). DySNet realizes this by the alternate stacking of dynamic stream block (DSB). This feature modeling network is modular and can be integrated with various operations (pooling, skip connections or other modules) to construct diverse DMIR architectures. This network achieves dynamic receptive fields and weights through two key innovations:

\textit{Adaptive Stream Basin (AdSB)} dynamically adjusts the window shapes of the computing unit according to the differences between input images (Fig.\ref{i2}-b), thereby adjusting the receptive fields of the model and enabling it to achieve adaptive dynamic search space. It simulates the process of stream flowing along the basin through a dynamic receptive fields \cite{fluid}, guiding the model's feature search space to adapt to the "basin" structure of the input images. Ultimately, AdSB eliminates interfering features by dynamically adjusting the search space, significantly reducing the calculation range of the feature relationships, enabling the model to focus more on potential feature combinations.

\textit{Dynamic Stream Attention (DySA)} calculates the weights according to the similarity of each element in the dynamic receptive fields provided by AdSB, and generates dynamic weights based on the point attention mechanism. These dynamically generated weights adaptively adjust the relationship search direction of the model (Fig.\ref{i2}-c), enabling the model to select the most accurate feature relationships in the search space. Ultimately, DySA enables the model to dynamically model feature relationships like a stream, improving the model's generalization ability.

Our contributions are summarized as follows: 1) We proposed the DySNet and DSB for the \textit{combinatorial explosion} problem in DMIR. This network achieves its adaptive capability through dynamically adjusting the receptive fields and weights of the model, enabling the model to eliminate interfering features and enhance its ability to discover potential feature relationships. 2) We proposed the AdSB module, which dynamically adjust the receptive field of each pixel in the image, thereby achieving a dynamic search space for relationships. 3) We proposed DySA, which dynamically change the search direction of feature relationships, thereby enhancing the modeling capability of complex spatial relationships during the registration process. 4) We instantiated DySNet into the architectures of Xmorpher \cite{xmorpher} and ModeT \cite{modet}, thereby creating DySNet-X and DySNet-M. Subsequent experiments revealed that the registration accuracies of both DySNet-X and DySNet-M were higher than those of the original networks.
\section{Related}
\label{sec:formatting}

\textbf{2.1 Deformable medical images registration.} Deformable medical image registration aims to align the anatomical structures of images by estimating non rigid transformations. Traditional optimization-based methods~\cite{thirion1998image,rueckert2002nonrigid,avants2008symmetric}rely on intensity similarity and smoothness constraints, but they are sensitive to initialization. Deep learning methods have gained popularity and be classified into the following three types: 1) Supervised methods require the use of real displacement fields or anatomical labels for supervision~\cite{krebs2017robust,krebs2017unsupervised}. 2) Unsupervised methods utilize transformation modules to minimize similarity losses and combine regularization ~\cite{voxelmorph,swin_voxelmorph,xmorpher,lku,transmorph,vit-v-net,sacb,adacs}. 3) Semi-supervised or self-supervised methods combine a small number of labels to enhance local anatomical consistency and modality-independent robustness~\cite{HU20181}. 

\textbf{2.2 Feature modeling in deformable medical images registration.} Feature modeling refers to the process by which the model transforms the original data into data that the machine understand and captures the relationships among the features~\cite{bengio2013representation}. Feature modeling in the existing registration networks can be classified into three categories. 1) Feature modeling based on CNNs~\cite{voxelmorph, lku} maintains the static receptive fields and weights learned by the basic unit convolution kernels~\cite{cnn} during the training and remains static during the test. Although it provides continuous feature representations that match medical image~\cite{deeplearning}, its static receptive fields and weights limit its adaptability to the complex and variable anatomical deformations in DMIR~\cite{bias}. 2) Feature modeling based on transformers~\cite{xmorpher, swin_voxelmorph,modet,transmorph} attempts to replace the convolution kernels with attention units to achieve dynamic modeling. However, their local partitioning mechanisms (patch~\cite{vit-v-net} or sliding window partition~\cite{swin_voxelmorph,xmorpher}) introduce artificial boundary effects, which disrupts the continuity~\cite{nnwnet}. 3) Other hybrid feature modeling methods alternate between these two modeling methods, which disrupts the smooth integration of local and global information~\cite{nnwnet}, resulting in feature inconsistency.

\textbf{2.3 Combinatorial explosion in feature modeling.} Combinatorial explosion refers to the phenomenon where the number of feature combinations increases exponentially with the dimension and interaction order during the feature modeling process, ultimately leading to a significant increase in model search complexity\cite{deepNET}. In the field of computer vision, as the number of pixels, channels of features grow, the number of feature combinations expands rapidly, thereby causing the problem of combinatorial explosion in tasks such as segmentation and registration\cite{deepNET}. To alleviate this issue, common methods include structured sparsity\cite{yuan2006model}, low-rank decomposition\cite{jaderberg2014speeding}, and attention mechanisms\cite{attention}. Although these methods have achieved practical results in different tasks, in applications such as registration where higher-order relationship modeling is required, combinatorial explosion remains a core challenge that limits performance and scalability.

\textbf{2.4 Dynamics in feature modeling.} Existing works have incorporated dynamics into the feature modeling. For instance, deformable convolution \cite{dcn,dcn2} dynamically adjust the sampling positions based on the input, enabling it to capture the geometric deformation features of the image. The same dynamic mechanism has been introduced into the vision Transformer (ViT \cite{vit}) to form DAT \cite{dat}. Although DAT achieves spatial sampling and attention aggregation through global key points, it is still limited by the fixed receptive fields input. SACB-Net \cite{sacb} adjusts the sampling and aggregation through feature clustering-based convolution kernels, but it is still difficult to achieve pixel level continuous variable receptive fields.

\section{Methodology}

\begin{figure*}[t] 
  \centering
  \includegraphics[width=\textwidth]{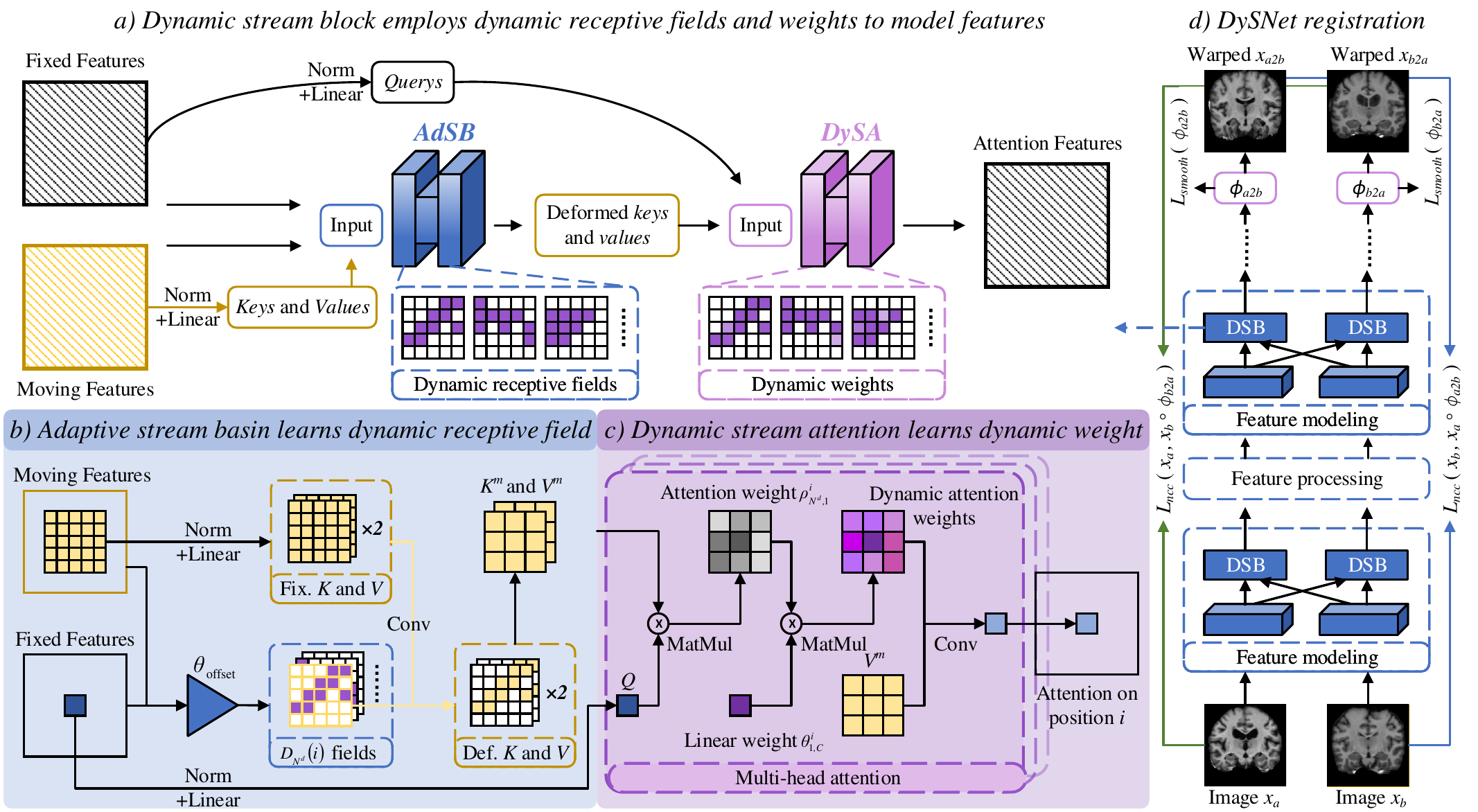}
  \caption{Overall architecture of our DySNet. a) Dynamic stream block generates queries, keys, and values according to fixed features and moving features, and combines the AdSB and the DySA to respectively construct dynamic receptive fields and dynamic weights, ultimately forming attention features. b) The AdSB learns the dynamic receptive fields. c) The DySA calculates dynamic weights of feature modeling. d) the general network architecture of DySNet.}
  \label{i3}
\end{figure*}

Our DySNet consists of two key modules (Fig.~\ref{i3}), AdSB and DySA, which jointly learn dynamic receptive fields and weights to adaptively model feature relationships. These modules are integrated into a bidirectional architecture that enhances accuracy and deformation smoothness in DMIR.

\subsection{Formulation}

\textbf{3.1.1 Problem formulation} In DMIR, the spatial domain $\Omega$ is composed of $H \times W$ pixel positions (H and W represent the height and width of the input image), where each position represents a feature point. For a given feature point $f_i$, the number of feature combinations in the image is proportional to the image resolution. Specifically, if we represent the average number of possible feature relationships for each pair of feature combinations as $\alpha$, then the size of the candidate feature combination relationship set for $f_i$ can be expressed as $c = \alpha \times H \times W - 1$, which grows linearly with the image resolution $H \times W$. Considering the entire image, which contains $N = H \times W$ features, and assuming independence of matching choices across features, the total number of possible feature combination relationships is
\begin{equation}
|\mathcal{H}| = c^N = (\alpha H W - 1)^{H W},
\end{equation}
indicating a exponential growth in the number of combinational relationships as image resolution ($H \times W$) increases. 

\textbf{3.1.2 Why dynamic methods reduce the number of feature combinations?} Each pixel position $i \in \Omega$, there is a static receptive fields $U_{N^{d}}(i)$ ($N^{d}$ is the size of static receptive fields). The static weights determine the set of feature relationships modeled in the receptive field for each pixel as $W(i)$, and $|W(i)| = m$ indicates the number of feature relationships considered for that pixel. Dynamic receptive fields learned by the AdSB module deform the static receptive fields $U_{N^{d}}(i)$ to a locally optimized neighborhood $D_{N^{d}}(i)$ (Eq.\eqref{eq:deformed_window}). This dynamic receptive fields effectively reduces the number of features in the static receptive field to approximately $|D_{N^{d}}(i)|$, where dynamic receptive fields $|D_{N^{d}}(i)|$ achieve the same ability of the static receptive fields $|U_{N^{d}}(i)|$ with fewer pixels ($|D_{N^{d}}(i)| < |U_{N^{d}}(i)|$). Then, the DySA module dynamically calculates the weights $\rho_{N^{d}, 1}^i(j)$ (Eq.\eqref{eq:softmax_attention}), based on the similarity of the features, reducing the number of feature relationships that each pixel needs to model in the dynamic receptive fields from $|W(i)|$ to $|A(i)|$ ($|A(i)| < |W(i)|$). Ultimately, the dynamic receptive fields and weights enable the model to reduce the number of feature combinations:
\begin{equation}
|D_{N^{d}}(i)| \times |A(i)| \ll |U_{N^{d}}(i)| \times |W(i)|.
\end{equation}

\subsection{AdSB learns dynamic receptive fields}

Our AdSB decomposes the receptive fields and weights through kernel decomposition to introduce dynamics subsequently. Then, the dynamic offset $\Delta i$ is determined through the offset prediction network to adjust the shape of the static receptive fields $U_{N^{d}}(i)$, and the dynamic receptive field is constructed for the operation of DySA.

\textbf{Kernel decomposition}  
Treat the basic kernel $K_{N^{d},\theta}(i)$ of model as a point-based attention mechanism to introduce two kinds of dynamics. The basic kernel of size $N^{d}$ and parameters $\theta_{N^{d},C}^i \in \mathbb{R}^{N^{d} \times C}$ are denoted as $K_{N^{d},\theta}(i)$, which be decomposed into receptive fields $U_{N^{d}}(i)$ and weights $\theta_{N^{d},C}^i$, and then weights is decomposed into spatial weights $\theta_{N^{d},1}^i \in \mathbb{R}^{N^{d} \times 1}$ and channel weights $\theta_{1,C}^i \in \mathbb{R}^{1 \times C}$:
\begin{equation}
\begin{cases}
K_{N^{d},\theta}(i) \to \{U_{N^{d}}(i), \theta_{N^{d},C}^i\}, \\
\theta_{N^{d},C}^i \to \{\theta_{N^{d},1}^i, \theta_{1,C}^i\}.
\end{cases}
\end{equation}
Then the receptive field is modified to achieve a dynamic by offset prediction. Subsequently, we propose to dynamically calculate the spatial weight $\rho_{N^{d},1}^i$ at position $i$ through the function $P(\cdot)$ based on the features of the current position and its neighborhood to achieve dynamic weights:
\begin{equation}
\rho_{N^{d},1}^i = P(f_i^a, f_j^b), \quad \forall j \in D_{N^{d}}(i),
\end{equation}
where $f_i^a$ and $f_j^b$ are features at position $i$ in the fixed image and its neighborhood in the moving image respectively, and $D_{N^{d}}(i)$ is the dynamic receptive field centered at $i$.

\textbf{Offset prediction constructs dynamic receptive fields}  
Specifically, given fixed image features $f^a$ and moving image features $f^b$, we first concatenate them along the channel dimension to form a fused feature representation $X = [f^a, f^b] \in \mathbb{R}^{B \times 2C \times H \times W}$, where $B$ is the batch size, $C$ is the number of channels, and $H,W$ denote spatial dimensions. Then, as shown in Fig.\ref{i3}-b an offset prediction network $\theta_{\mathrm{offset}}$ predicts a set of dynamic offsets for each pixel in the receptive fields $U_{N^{d}}(i)$ at every location $i$: $\Delta i = \theta_{\mathrm{offset}}(X) \in \mathbb{R}^{B \times d|U_{N^{d}}| \times H \times W}$. Then reshape the offsets as: $\Delta i \in \mathbb{R}^{B \times |U_{N^{d}}| \times d \times H \times W}$, corresponding to the offsets of all sampled points at every location. The dynamic receptive fields $D_{N^{d}}(i)$ (Fig.\ref{i3}-b) is then defined as:
\begin{equation}
D_{N^{d}}(i) = U_{N^{d}}(i) + \Delta i,
\label{eq:deformed_window}
\end{equation}
where $U_{N^{d}}(i)$ denotes the static receptive fields. Based on the continuity of the offsets, we apply the interpolation function $I(\cdot)$ to the key and value features $K, V \in \mathbb{R}^{B \times C \times H \times W}$ to obtain the deformed keys and values:
\begin{equation}
K_j^m = I\left(K, D_{N^{d}}(i)_j\right), \quad V_j^m = I\left(V, \mathcal{D}_{N^{d}}(i)_j\right),
\end{equation}
where $j=1,...,|D_{N^{d}}(i)|$ indexes each pixel in the field. By learning dynamic offsets $\Delta i$, AdSB allows the receptive fields to flexibly deform and focus on relevant features.

\textbf{Discussion of properties}: 1) Decoupling the structure of the kernel facilitates the introduction of dynamics. 2) AdSB learns the deformed receptive fields reduce interference features and improves the efficiency and accuracy of feature modeling. The interpolation sampling ensures the continuity and differentiability of the sampling, which is conducive to stable training and smooth deformation.

\subsection{DySA learns dynamic weights}

Our DySA module calculates the attention to adjust the spatial weights of the dynamic receptive field obtained from AdSB: $\rho_{N^{d},1}^i = P(Q_i, K_j^m)$. Given a query feature $Q_i$ from $f^a$ at position $i$ and deformed keys $K_j^m$ from $f^b$ in $U_{N^{d}}(i)$, the dot product similarity is computed as $e_{ij} = \frac{Q_i^\top K_j^m}{\sqrt{c}}$, where $c$ is the dimensionality of each attention head, used to scale the dot product. The attention $\rho_{N^{d},1}^i(j)$ are then obtained by applying a softmax normalization over all sampled points in the dynamic receptive fields:
\begin{equation}
\rho_{N^{d},1}^i(j) = \frac{\exp(e_{ij})}{\sum_{j=1}^{|D_{N^{d}}(i)|} \exp(e_{ij})}, \quad j=1,...,|D_{N^{d}}(i)|,
\label{eq:softmax_attention}
\end{equation}
\textbf{Discussion of properties}: Point attention calculation of the DySA module dynamically weight and aggregate features in the dynamic receptive fields, thereby enhancing the model's adaptability to local deformations. Moreover, the softmax based normalization ensures that the attention scores form a proper probability distribution, stabilizing training and enabling interpretable attention maps that highlight critical regions contributing to the registration.

\begin{table*}[htbp]
  \caption{The evaluations on a) 3D cardiac structures, b) 3D brain tissues and c) 2D brain tissues demonstrates our strong ability in modeling feature relationships. The column $|J_{\phi}| < 0\% \downarrow$ shows the maximal Jacobian negative volume among tasks as a conservative metric. “AVG” is the arithmetic mean of three DSC scores (columns a, b, c) in each row. “*” means invalid values due to registration failure. The best and second-best DSC scores per column highlight state-of-the-art performance of our method.}
  \label{tab:merged_seg4}
  \centering
  \begin{tabular}{@{}llccccccl@{}}
    \toprule
    Method & Net. & \multicolumn{2}{c}{a) 3D Cardiac CT} & \multicolumn{2}{c}{b) 3D Brain MRI} & \multicolumn{2}{c}{c) 2D Brain MRI} & AVG \\
           &         & DSC \%  & $|J_{\phi}|\% $ & DSC \%  & $|J_{\phi}|\% $ & DSC \%  & $|J_{\phi}|\% $ & DSC  \\
    \midrule
    Initial     & -       & $62.5_{\pm 8.7}$  & -          & $65.5_{\pm 3.2}$  & -          & $62.1_{\pm 6.9}$  & -         & 63.4 \\
    LKU-Net~\cite{lku}        & C     & $64.0_{\pm 12.8}$  & $0.10_{\pm 0.03}$ & $73.6_{\pm 8.0}$  & $0.10_{\pm 0.02}$ & $69.5_{\pm 12.1}$ & \textcolor{red}{$0.01_{\pm 0.01}$} & 69.0 \\
    VoxelMorph~\cite{voxelmorph}     & C     & $77.0_{\pm 13.4}$  & \textcolor{blue}{$0.08_{\pm 0.04}$} & $75.9_{\pm 7.8}$  & \textcolor{blue}{$0.05_{\pm 0.02}$} & $78.6_{\pm 11.6}$ & \textcolor{blue}{$0.17_{\pm 0.12}$} & 77.2 \\
    SACB~\cite{sacb}      &  C   & $83.0_{\pm 12.8}$ & $1.19_{\pm 0.30}$  & \textcolor{blue}{$78.9_{\pm 8.0}$} & $0.40_{\pm 0.06}$ & \textcolor{blue}{$82.5_{\pm 11.8}$} & $0.92_{\pm 0.27}$ & \textcolor{blue}{81.5} \\
    ViT-V-Net~\cite{vit-v-net}       & C+T & $73.5_{\pm 13.3}$  & $2.88_{\pm 0.65}$ & $76.8_{\pm 7.9}$  & $0.60_{\pm 0.09}$      & $49.1^{*}_{\pm 10.1}$ & $0.79_{\pm 0.26}$ & 66.5 \\ 
    TransMorph~\cite{transmorph}      & C+T & $69.0_{\pm 13.2}$  & $1.94_{\pm 0.25}$ & $71.7_{\pm 14.3}$  & $0.57_{\pm 0.10}$ & $82.3_{\pm 11.9}$ & $0.94_{\pm 0.30}$ & 74.3 \\
    Swin-VM~\cite{swin_voxelmorph}        & T  & $62.7^{*}_{\pm 8.0}$  & $0^{*}$ & $65.2^{*}_{\pm 7.3}$  & $0^{*}$ & $82.2_{\pm 11.8}$ & $0.77_{\pm 0.25}$ & 70.0 \\
    \midrule
    XMorpher~\cite{xmorpher}       & T   & $72.4_{\pm 13.4}$  & \textcolor{red}{$0.03_{\pm 0.02}$} & $71.2_{\pm 7.5}$  & \textcolor{red}{$0.03_{\pm 0.01}$} & $76.5_{\pm 11.5}$ &$0.80_{\pm 0.06}$ & 73.4 \\
    DySNet-X  & T   & $74.5_{(+2.1)\pm 13.2}$ & $0.98_{\pm 0.88}$ & $77.8_{(+6.6)\pm 8.0}$  & $0.33_{\pm 0.06}$ & \textcolor{red}{$83.0_{(+6.5)\pm 11.9}$} & $0.90_{\pm 0.28}$ & 78.4 \\
    \midrule
    ModeT~\cite{modet}      & T   & \textcolor{blue}{$83.6_{\pm 12.8}$} & $0.73_{\pm 0.21}$  &   $77.4_{\pm 8.0}$ & $0.22_{\pm 0.03}$  & $81.9_{\pm 11.8}$ & $0.81_{\pm 0.25}$ & 81.0 \\
    DySNet-M  & T   & \textcolor{red}{$84.1_{(+0.5)\pm 12.6}$} & $0.76_{\pm 0.27}$ & \textcolor{red}{$79.7_{(+2.3)\pm 8.0}$}  & $0.31_{\pm 0.06}$ & $82.2_{(+0.3)\pm 11.8}$ &  $0.79_{\pm 0.24}$ & \textcolor{red}{82.0} \\
    \bottomrule
  \end{tabular}
\end{table*}

\subsection{DSB constructs dynamic kernel}

Our DSB combines the dynamic receptive fields from AdSB and dynamic weights from DySA to form a unified dynamic kernel for feature aggregation.

\textbf{Dynamic kernel construction} The dynamic kernel at position $i$ is then obtained by fusing the attention weights $\rho_{N^{d},1}^i(j)$ with the learnable channel weight $\theta_{1,C}^i$ in dynamic receptive fields $D_{N^{d}}(i)$, as show in Fig.\ref{i3}-c:
\begin{equation}
\begin{cases}
\{\rho_{N^{d},1}^i, \theta_{1,C}^i\} \to \theta_{N^{d},C}^i, \\
\{D_{N^{d}}(i), \theta_{N^{d},C}^i\} \to K_{N^{d},\theta}(i).
\end{cases}
\end{equation}
\textbf{Feature aggregation} Therefore, this kernel $K_{N^{d},\theta}(i)$ is used as to extract the features $A_i$ on position $i$ corresponding to $N^{d}$ field in deformed value features $V_j^m$:
\begin{equation}
A_i = \sum_{j=1}^{|D_{N^{d}}|} \theta_{N^{d},C}^i(j) \cdot V_j^m,
\label{eq:weighted_sum}
\end{equation}
achieving attention values at position $i$ in one head by computing weighted sums. The results from all heads are then summed to produce the final attention values, allowing the model to capture richer contextual information.

\textbf{Discussion of properties}: The dynamic kernel adaptively modulates spatial receptive fields and weights, enabling feature aggregation with enhanced representational capacity. This dynamic adaptation facilitates improved feature modeling and robustness to spatial variations.

\subsection{Details of DySNet}
DMIR is formulated as a bijective mapping problem \cite{koopman1982variables} between the shared anatomical content of two images, $x_a$ and $x_b$. This formulation imposes a structural prior that the registration function should be symmetric, meaning the mapping from $x_a$ to $x_b$ and from $x_b$ to $x_a$ are inverse to each other. Enforcing this symmetry ensures consistent spatial correspondence between shared structures. To incorporate this prior, we adopt a bidirectional registration framework along with a symmetric loss function. So in Fig.\ref{i3}-d, DySNet is achieved by stacking multiple DSBs (F) symmetrically. The definition of the DySNet registration is as follows, $F(x_a, x_b; \theta) = F_T^*\Big(\cdots F_2^*\big(F_1^*, (F_1^*)^\top\big); \theta \Big)$ ,where $(F_1^*)^\top$ denotes the symmetric counterpart with swapped inputs. We incorporated our DySNet into the architecture of Xmorpher \cite{xmorpher}, creating \textbf{DySNet-X} by replacing its CAT blocks. We also integrated ModeT's \cite{modet} architecture by replacing its ModeT module to form \textbf{DySNet-M}. \textbf{Symmetric loss function} is formulated as:
\begin{equation}
\mathcal{L}_{bireg} = \mathcal{L}_{reg}(x_b, x_{a 2 b}, \phi_{a 2 b}) + \mathcal{L}_{reg}(x_a, x_{b 2 a}, \phi_{b 2 a}),
\label{eq:bidirectional_loss}
\end{equation}
where $x_{a 2 b} = x_a \circ \phi_{a 2 b}$ and $x_{b 2 a} = x_b \circ \phi_{b 2 a}$ are images warped by the symmetry deformation fields respectively. Following Xmorpher, $\mathcal{L}_{reg}$ is a deformable registration loss composed of smoothness loss $\mathcal{L}_{smo}$ (Jacobian matrix), which preserves deformation topology, and similarity loss $\mathcal{L}_{sim}$ (mean dice similarity coefficients ), which aligns similar regions.

\section{Experiments}

\subsection{Experimental protocol}
\textbf{\textit{Materials:}} We evaluate the excellent DMIR performance of our DySNet on three tasks with different imaging modalities and features. 1) 3D Cardiac structure registration evaluate our method on seven large cardiac structures on computed tomography (CT) images. It fuses three publicly available datasets, including the MM-WHS~\cite{MM-WHS} (20 images with cardiac structure labels and 40 unlabeled images), ASOCA~\cite{ASOCA}, (60 unlabeled images), and CAT08\cite{CAT08} (32 images with cardiac structure labels). Therefore, a total of 100 unlabeled images are enrolled in our evaluation for 9,900 image pairs as training dataset. The 52 labeled images compose 2,652 image pairs as testing dataset. Cardiac regions in the images are cropped and resampled to 144 × 144 × 128. 2) 3D Brain tissues registration evaluate our framework on 28 small brain tissues on T1 magnetic resonance (MR) images. It collects 837 T1 brain MR images from the PPMI~\cite{ppmi} database and composes 699,732 image pairs as our training dataset. Another dataset, CANDI~\cite{candi}, with 103 T1 brain MR images with brain tissue labels is also enrolled in evaluation for 10,506 image pairs as the testing dataset. The brain regions are extracted by the HD-BET~\cite{HD-BET}, and cropped and resampled to 160×160×128. 3) 2D Brain tissues registration on OASIS-1. We divided the official dataset into 361 as the training set and 53 as the test set. Which means it composes 129,960 image pairs as training dataset and 2,756 image pairs as the testing dataset. The  2D data provided by the authorities are cropped to 160×160.

\textbf{\textit{Comparisons:}} To benchmark our framework, we take 8 works including CNN-based methods~\cite{voxelmorph,lku,sacb}, Transformer-based methods~\cite{xmorpher,modet,swin_voxelmorph}, and CNN-Transformer-fused methods~\cite{transmorph,vit-v-net} in DMIR. Therefore, the superiority of our module will be demonstrated.

\textbf{\textit{Implementation and evaluation metrics:}} We conducted a comprehensive evaluation of our registration performance from alignment accuracy and smoothness. We used the Dice similarity coefficient (DSC $\pm_{std}[\%]$) and the percentage of Jacobian matrix $<0$ ($|J_{\phi}| < 0 \pm_{std}[\%]$) to assess the alignment accuracy and smoothness. Our DySNet were conducted on NVIDIA RTX 6000 with 24\,GB of memory, using PyTorch~\cite{pytorch}. We used AdamW \cite{adamw} as the optimizer with an initial learning rate of $10^{-4}$. To conduct a fair comparison, all the methods employed in our experiment either adopted the same hyperparameter settings or adhered to the experimental settings provided in the original papers.

\begin{figure*}[t]
  \centering
   \includegraphics[width=1\linewidth]{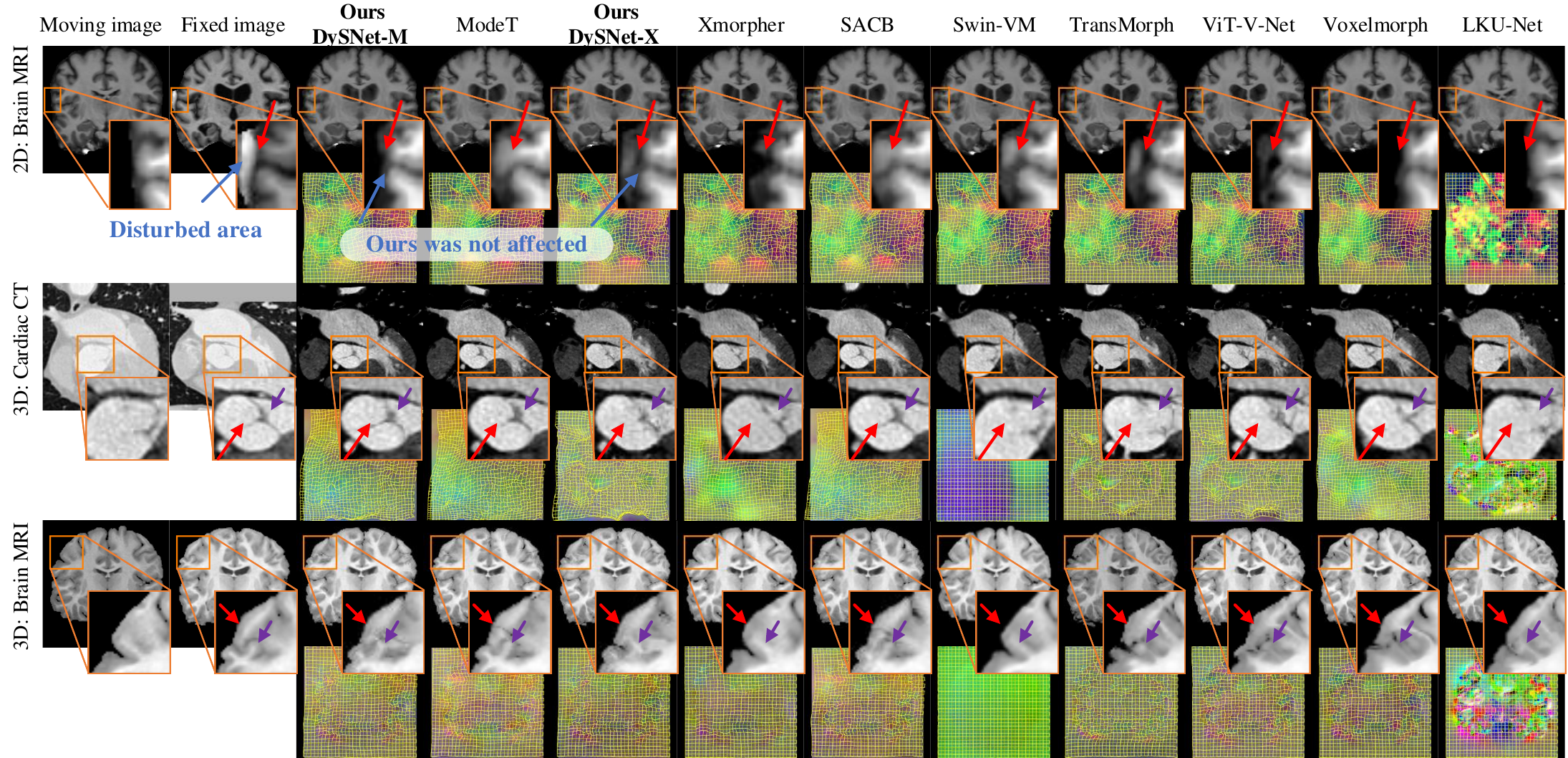}
   \caption{This figure presents a qualitative comparison of various registration methods between dynamic images and fixed images. It includes our proposed DySNet, as well as mainstream registration networks. The results show that our method outperforms others in maintaining details and structural consistency and it also significantly improves upon methods with the same network structure (ModeT and Xmorpher). “*" mean the values are invalid because the methods registration failed.}
   \label{i4}
\end{figure*}

\subsection{Quantitative analysis}

We evaluate the performance of our DySNet on three registration tasks involving 3D cardiac CT, 3D and 2D brain MRI, comparing with eight state-of-the-art methods. The results in Tab.\ref{tab:merged_seg4} highlight three key observations:

\textbf{\textit{Superior registration accuracy across diverse anatomical scales and modalities:}} Our DySNet demonstrates superior registration accuracy across diverse anatomical scales and imaging modalities, achieving an average DSC of 82.0\%, significantly outperforming other state-of-the-art methods. Specifically, DySNet attains the highest DSC of 84.1\% on 3D cardiac CT, surpassing traditional convolutional methods like VoxelMorph by over 7 percentage points. On 3D brain MRI, DySNet achieves 79.7\%. For 2D brain MRI, DySNet reaches the top DSC of 83.0\%. In contrast, convolutional methods such as VoxelMorph~\cite{voxelmorph} and LKU-Net~\cite{lku} show moderate accuracy, and hybrid methods like ViT-V-Net and TransMorph~\cite{transmorph} suffer from unstable performance and occasional registration failures, notably reducing DSC in some tasks.

\textbf{\textit{Robust and plausible deformation fields:}} Our DySNet maintains low Jacobian negative volume values, indicating physically plausible and topology-preserving transformations. For example, DySNet records 0.79\% negative Jacobian volume on 2D brain MRI. VoxelMorph~\cite{voxelmorph} also exhibits relatively low negative Jacobian volumes (0.08\% on 3D cardiac CT, 0.05\% on 3D brain MRI, and 0.17\% on 2D brain MRI), reflecting stable deformations. By comparison, methods like SACB~\cite{sacb}, ViT-V-Net, and TransMorph~\cite{transmorph} show notably higher negative Jacobian volumes (up to 2.88\%), sometimes accompanied by registration failures. Although XMorpher~\cite{xmorpher} achieves the lowest negative Jacobian volume of 0.03\%, its registration accuracy remains substantially lower than DySNet’s. These results demonstrate that DySNet achieves an optimal balance between high registration accuracy and smooth deformation fields, confirming its superior performance and robustness.

\textbf{\textit{Improvements in two network architectures:}} Our method substantially enhances accuracy on architectures such as XMorpher~\cite{xmorpher} and ModeT~\cite{modet}, with especially notable improvements on fine anatomical structures like 3D brain tissues (+6.6\% and +2.9\%). This is attributed to the integration of dynamic receptive fields and point attention, which enables more precise, spatially adaptive feature aggregation. Furthermore, our experiments have shown that networks employing the pyramid architecture (such as ModeT~\cite{modet} and SACB~\cite{sacb}) demonstrate outstanding performance by leveraging multi-scale contextual information.

\subsection{Qualitative evaluation}

In Fig.~\ref{i4}, we show typical cases on the three DMIR tasks in this experiment and our DySNet has high accuracy and few folds in the whole image. It has two observations:

\textbf{\textit{Accurate registration for fine structures:}} For brain tissues, DySNet demonstrates outstanding alignment performance. Especially within the 2D magnified area, where the interfering area is the remaining brain structure, DySNet remains unaffected and achieves accurate registration. In the 3D magnified area, our method achieves smooth registration, while in SACB~\cite{sacb} and Modet~\cite{modet}, there are split situations. This is mainly due to the dynamic mechanism adopted by DySNet, which can capture more detailed spatial relationships and avoid the disruption of continuity, thereby achieving precise alignment at the detail level. For cardiac structures, DySNet also generate fine deformations at the boundaries while maintaining continuous context (visible in the magnified area). In contrast, other methods (TransMorph~\cite{transmorph} and VoxelMorph~\cite{voxelmorph}) exhibit obvious mixing phenomena at the boundary of the magnified area.

\textbf{\textit{Smooth registration preserves the reality:}} By visualizing the deformation vector field (DVF) in a grid form, DySNet demonstrates reasonable smoothness, effectively ensuring the authenticity and rationality of the deformed images. For the heart structure, the artificially divided spatial block-like structure in the Transformer method is prone to causing distortion in spatial transformation, and even fails in registration in Swin-VM~\cite{swin_voxelmorph}. ViT-V-Net, due to its use of the patch-based attention mechanism of ViT, performs poorly on the deformation of heart structure. Xmorpher~\cite{sacb} is smooth but has lower registration accuracy. In contrast, DySNet, by calculating dynamic weights, combines strong smoothing for large scale heart structures and balanced registration performance for small scale brain tissue.

\begin{figure}[t]
  \centering
   \includegraphics[width=1\linewidth]{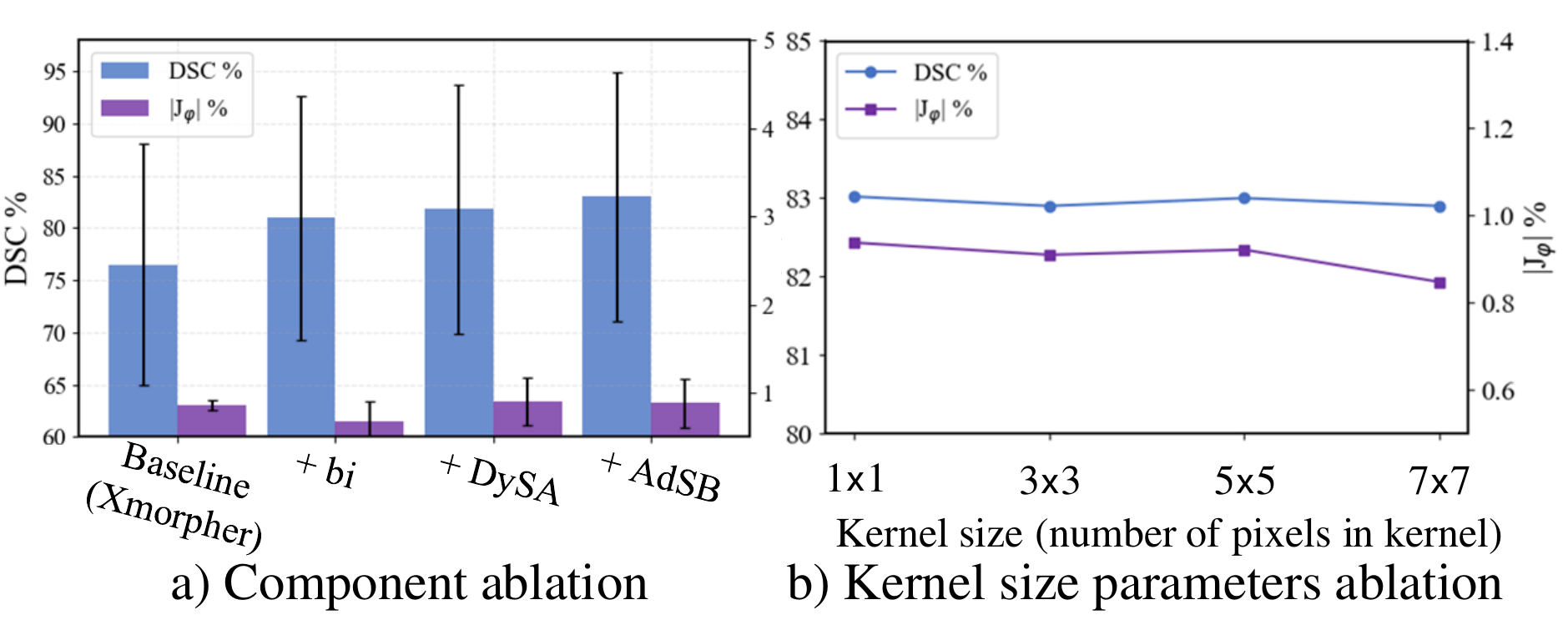}
   \caption{a) Ablation results of our DySNet-X on the 2D Brain MRI set. Add the bidirectional registration (bi), DySA and AdSB modules in sequence. b) Hyper-parameter ablation of kernel sizes.}
   \label{i5}
\end{figure}

\subsection{Ablation study and model analysis}

\textbf{4.4.1 Component ablation:} Ablation study of DySNet (Fig.~\ref{i5}-a) highlights the impact of each component. The baseline Xmorpher achieves a DSC of 76.5\% with a regularization error ($|J_{\phi}|$) of 0.80\%, showing moderate performance. Adding bidirectional registration (bi) significantly lowers the regularization error to 0.67\% and improves DSC to 81.0\%, indicating more stable deformations. Incorporating the DySA module further boosts DSC to around 82\% while maintaining low regularization error, enhancing semantic matching. Finally, integrating the AdSB module raises DSC to 83.0\% with a controlled regularization error of 0.88\%, achieving the best accuracy. These results confirm that jointly modeling dynamic receptive fields and weights effectively improves DMIR performance.

\textbf{4.4.2 Hyper-parameter ablation:} Fig.\ref{i5}-b illustrates the impact of varying the kernel size on our DySNet. Notably, although a larger kernel is used, the model still maintains excellent performance. For example, the mean DSC ranges from 82.9\% to 83.02\%, and the mean Jacobian determinant \(|J_{\phi}|\) varies between 0.848 and 0.938. These results show the model’s generalization capabilities learning in our dynamic method, proving the dynamic attention mechanism captures key feature relationships without large spatial contexts. This aligns with our goal of improving DMIR generalization through dynamic adjustment of receptive fields and weights.

\begin{figure}[t]
  \centering
   \includegraphics[width=1\linewidth]{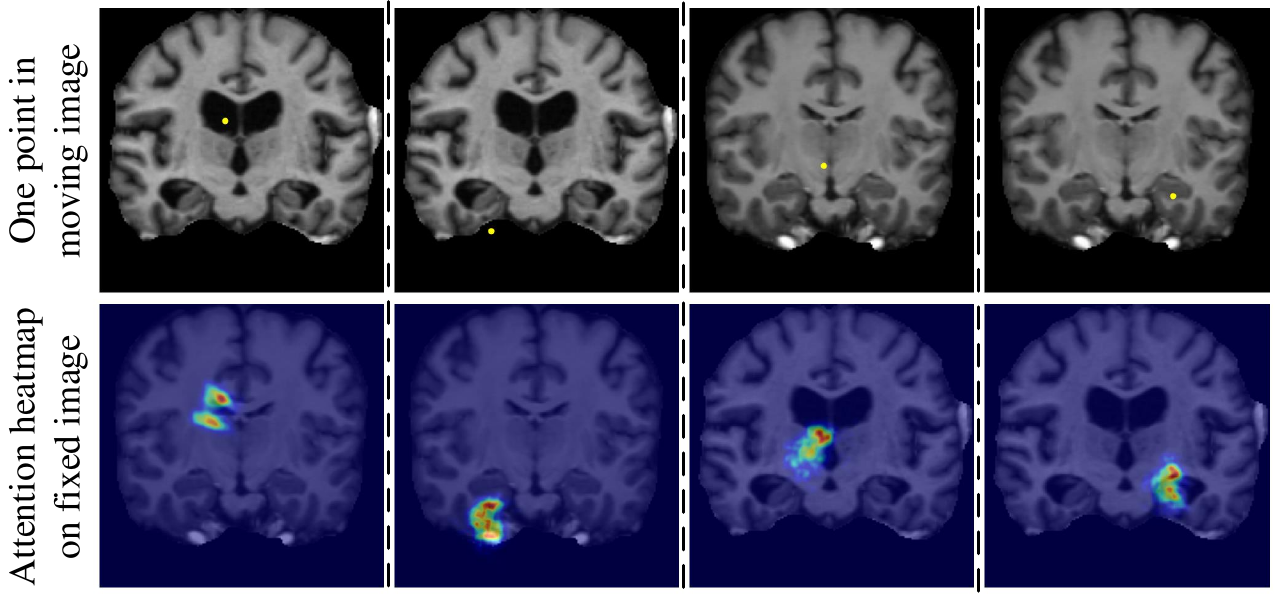}
   \caption{This figure illustrates the dynamic changes exhibited by our DySNet-X when performing feature modeling at different positions (as indicated by the yellow dots). The first row shows the dynamic image and the given position point and the second row presents the heatmap of the attention area}
   \label{i6}
\end{figure}

\textbf{4.4.3 Attention heatmaps analysis:} We selected representative points with notable displacements for visualization. As shown in Fig.\ref{i6} the attention heatmaps produced by DySA effectively highlight highly responsive regions in the fixed image, demonstrating the model’s ability to dynamically adjust both receptive fields and attention weights. This visualization provides clear evidence that the model accurately localizes features.

\textbf{4.4.4 Computation efficiency:} Registration efficiency varies little with kernel size, prompting investigation into model complexity as the kernel size changes. The results: \begin{center}
  \includegraphics[width=0.48\textwidth]{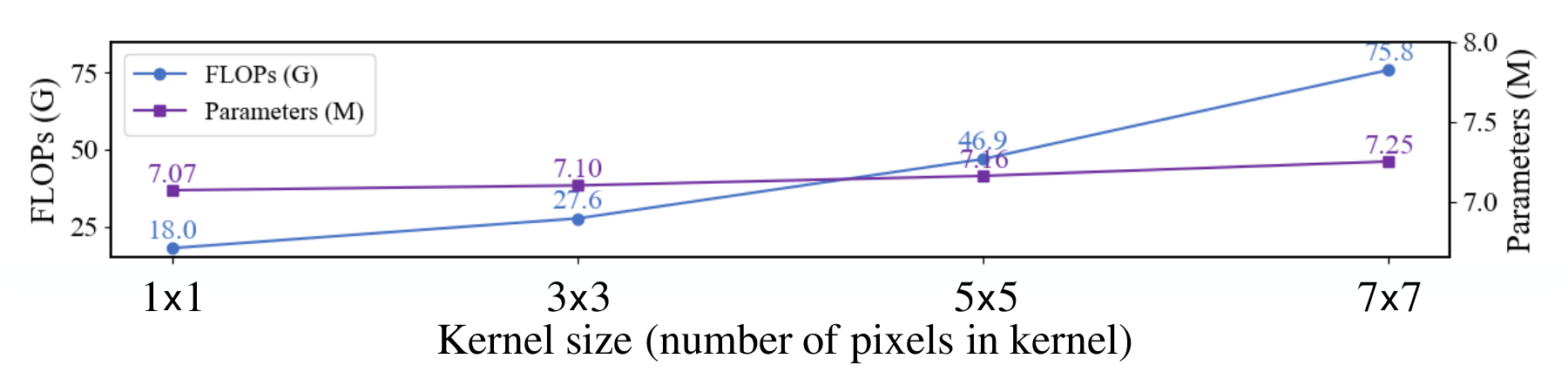}
\end{center} FLOPs increase significantly with kernel size, indicating higher computational cost, while parameter count remains nearly constant, reflecting stable model size. This indicates that kernel size mainly affects computation but has limited impact on model parameters, confirming the model's efficiency even at the smallest kernel size due to its dynamic.

\section{Conclusion}
We propose DySNet, a novel network that effectively improves the accuracy of DMIR by adaptively adjusting both receptive fields and weights. Extensive experimental results demonstrate that DySNet consistently achieves superior registration accuracy and deformation smoothness across diverse datasets, validating its robustness and generalizability. In the future we envision extending DySNet to a broader range of medical image analysis tasks, thereby unlocking further potential in applications.

{
    \small
    \bibliographystyle{ieeenat_fullname}
    \bibliography{main}

@article{adamw,
  title={Decoupled weight decay regularization},
  author={Loshchilov, Ilya and Hutter, Frank},
  journal={arXiv preprint arXiv:1711.05101},
  year={2017}
}

@article{pytorch,
  title={Pytorch: An imperative style, high-performance deep learning library},
  author={Paszke, Adam and Gross, Sam and Massa, Francisco and Lerer, Adam and Bradbury, James and Chanan, Gregory and Killeen, Trevor and Lin, Zeming and Gimelshein, Natalia and Antiga, Luca and others},
  journal={Advances in neural information processing systems},
  volume={32},
  year={2019}
}

@article{HD-BET,
  title={Automated brain extraction of multisequence MRI using artificial neural networks},
  author={Isensee, Fabian and Schell, Marianne and Pflueger, Irada and Brugnara, Gianluca and Bonekamp, David and Neuberger, Ulf and Wick, Antje and Schlemmer, Heinz-Peter and Heiland, Sabine and Wick, Wolfgang and others},
  journal={Human brain mapping},
  volume={40},
  number={17},
  pages={4952--4964},
  year={2019},
  publisher={Wiley Online Library}
}

@article{ppmi,
  title={The Parkinson progression marker initiative (PPMI)},
  author={Marek, Kenneth and Jennings, Danna and Lasch, Shirley and Siderowf, Andrew and Tanner, Caroline and Simuni, Tanya and Coffey, Chris and Kieburtz, Karl and Flagg, Emily and Chowdhury, Sohini and others},
  journal={Progress in neurobiology},
  volume={95},
  number={4},
  pages={629--635},
  year={2011},
  publisher={Elsevier}
}

@article{candi,
  title={CANDIShare: a resource for pediatric neuroimaging data},
  author={Kennedy, David N and Haselgrove, Christian and Hodge, Steven M and Rane, Pallavi S and Makris, Nikos and Frazier, Jean A},
  journal={Neuroinformatics},
  volume={10},
  number={3},
  pages={319--322},
  year={2012},
  publisher={Springer}
}

@article{CAT08,
  title={An automatic multi-tissue human fetal brain segmentation benchmark using the fetal tissue annotation dataset},
  author={Payette, Kelly and de Dumast, Priscille and Kebiri, Hamza and Ezhov, Ivan and Paetzold, Johannes C and Shit, Suprosanna and Iqbal, Asim and Khan, Romesa and Kottke, Raimund and Grehten, Patrice and others},
  journal={Scientific data},
  volume={8},
  number={1},
  pages={167},
  year={2021},
  publisher={Nature Publishing Group UK London}
}

@article{ASOCA,
  title={Automated segmentation of normal and diseased coronary arteries--the asoca challenge},
  author={Gharleghi, Ramtin and Adikari, Dona and Ellenberger, Katy and Ooi, Sze-Yuan and Ellis, Chris and Chen, Chung-Ming and Gao, Ruochen and He, Yuting and Hussain, Raabid and Lee, Chia-Yen and others},
  journal={Computerized Medical Imaging and Graphics},
  volume={97},
  pages={102049},
  year={2022},
  publisher={Elsevier}
}

@article{MM-WHS,
  title={Evaluation of algorithms for multi-modality whole heart segmentation: an open-access grand challenge},
  author={Zhuang, Xiahai and Li, Lei and Payer, Christian and {\v{S}}tern, Darko and Urschler, Martin and Heinrich, Mattias P and Oster, Julien and Wang, Chunliang and Smedby, {\"O}rjan and Bian, Cheng and others},
  journal={Medical image analysis},
  volume={58},
  pages={101537},
  year={2019},
  publisher={Elsevier}
}

@article{fluid,
  title={Fluid mechanics, 5th version},
  author={Kundu, Pijush K and Cohen, Ira M and Dowling, David R},
  journal={Academic, Berlin},
  year={2012}
}

@article{deepNET,
  title={Deep nets: What have they ever done for vision?},
  author={Yuille, Alan L and Liu, Chenxi},
  journal={International Journal of Computer Vision},
  volume={129},
  number={3},
  pages={781--802},
  year={2021},
  publisher={Springer}
}

@article{DMIRsurvey,
  title={Deformable medical image registration: A survey},
  author={Sotiras, Aristeidis and Davatzikos, Christos and Paragios, Nikos},
  journal={IEEE transactions on medical imaging},
  volume={32},
  number={7},
  pages={1153--1190},
  year={2013},
  publisher={IEEE}
}

@article{yosinski2014transferable,
  title={How transferable are features in deep neural networks?},
  author={Yosinski, Jason and Clune, Jeff and Bengio, Yoshua and Lipson, Hod},
  journal={Advances in neural information processing systems},
  volume={27},
  year={2014}
}

@article{koopman1982variables,
  title={Variables and the bijection principle},
  author={Koopman, Hilda and Sportiche, Dominique},
  year={1982},
  publisher={Walter de Gruyter, Berlin/New York Berlin, New York}
}

@inproceedings{nonlocal,
  title={Non-local neural networks},
  author={Wang, Xiaolong and Girshick, Ross and Gupta, Abhinav and He, Kaiming},
  booktitle={Proceedings of the IEEE conference on computer vision and pattern recognition},
  pages={7794--7803},
  year={2018}
}

@inproceedings{adacs,
  title={Adaptive Correspondence Scoring for Unsupervised Medical Image Registration},
  author={Zhang, Xiaoran and Stendahl, John C and Staib, Lawrence H and Sinusas, Albert J and Wong, Alex and Duncan, James S},
  booktitle={European Conference on Computer Vision},
  pages={76--92},
  year={2024},
  organization={Springer}
}

@article{rueckert2002nonrigid,
  title={Nonrigid registration using free-form deformations: application to breast MR images},
  author={Rueckert, Daniel and Sonoda, Luke I and Hayes, Carmel and Hill, Derek LG and Leach, Martin O and Hawkes, David J},
  journal={IEEE transactions on medical imaging},
  volume={18},
  number={8},
  pages={712--721},
  year={2002},
  publisher={IEEE}
}

@inproceedings{rohe2017svf,
  title={SVF-Net: learning deformable image registration using shape matching},
  author={Roh{\'e}, Marc-Michel and Datar, Manasi and Heimann, Tobias and Sermesant, Maxime and Pennec, Xavier},
  booktitle={International conference on medical image computing and computer-assisted intervention},
  pages={266--274},
  year={2017},
  organization={Springer}
}

@article{avants2008symmetric,
  title={Symmetric diffeomorphic image registration with cross-correlation: evaluating automated labeling of elderly and neurodegenerative brain},
  author={Avants, Brian B and Epstein, Charles L and Grossman, Murray and Gee, James C},
  journal={Medical image analysis},
  volume={12},
  number={1},
  pages={26--41},
  year={2008},
  publisher={Elsevier}
}

@article{thirion1998image,
  title={Image matching as a diffusion process: an analogy with Maxwell's demons},
  author={Thirion, J-P},
  journal={Medical image analysis},
  volume={2},
  number={3},
  pages={243--260},
  year={1998},
  publisher={Elsevier}
}

@inproceedings{sacb,
  title={SACB-Net: Spatial-awareness Convolutions for Medical Image Registration},
  author={Cheng, Xinxing and Zhang, Tianyang and Lu, Wenqi and Meng, Qingjie and Frangi, Alejandro F and Duan, Jinming},
  booktitle={Proceedings of the Computer Vision and Pattern Recognition Conference},
  pages={5227--5237},
  year={2025}
}

@inproceedings{krebs2017unsupervised,
  title={Unsupervised probabilistic deformation modeling for robust diffeomorphic registration},
  author={Krebs, Julian and Mou, Wen and Jiang, Sheng and Navab, Nassir},
  booktitle={International Conference on Medical Image Computing and Computer-Assisted Intervention},
  pages={729--737},
  year={2017},
  organization={Springer}
}

@article{voxelmorph,
  title={Voxelmorph: a learning framework for deformable medical image registration},
  author={Balakrishnan, Guha and Zhao, Amy and Sabuncu, Mert R and Guttag, John and Dalca, Adrian V},
  journal={IEEE transactions on medical imaging},
  volume={38},
  number={8},
  pages={1788--1800},
  year={2019},
  publisher={IEEE}
}

@article{luo2016understanding,
  title={Understanding the effective receptive field in deep convolutional neural networks},
  author={Luo, Wenjie and Li, Yujia and Urtasun, Raquel and Zemel, Richard},
  journal={Advances in neural information processing systems},
  volume={29},
  year={2016}
}

@inproceedings{he2016deep,
  title={Deep residual learning for image recognition},
  author={He, Kaiming and Zhang, Xiangyu and Ren, Shaoqing and Sun, Jian},
  booktitle={Proceedings of the IEEE conference on computer vision and pattern recognition},
  pages={770--778},
  year={2016}
}

@inproceedings{unet,
  title={U-net: Convolutional networks for biomedical image segmentation},
  author={Ronneberger, Olaf and Fischer, Philipp and Brox, Thomas},
  booktitle={Medical image computing and computer-assisted intervention--MICCAI 2015: 18th international conference, Munich, Germany, October 5-9, 2015, proceedings, part III 18},
  pages={234--241},
  year={2015},
  organization={Springer}
}

@inproceedings{lku,
  title={U-net vs transformer: Is u-net outdated in medical image registration?},
  author={Jia, Xi and Bartlett, Joseph and Zhang, Tianyang and Lu, Wenqi and Qiu, Zhaowen and Duan, Jinming},
  booktitle={International Workshop on Machine Learning in Medical Imaging},
  pages={151--160},
  year={2022},
  organization={Springer}
}

@inproceedings{swin_transformer,
  title={Swin transformer: Hierarchical vision transformer using shifted windows},
  author={Liu, Ze and Lin, Yutong and Cao, Yue and Hu, Han and Wei, Yixuan and Zhang, Zheng and Lin, Stephen and Guo, Baining},
  booktitle={Proceedings of the IEEE/CVF international conference on computer vision},
  pages={10012--10022},
  year={2021}
}

@inproceedings{dat,
  title={Vision transformer with deformable attention},
  author={Xia, Zhuofan and Pan, Xuran and Song, Shiji and Li, Li Erran and Huang, Gao},
  booktitle={Proceedings of the IEEE/CVF conference on computer vision and pattern recognition},
  pages={4794--4803},
  year={2022}
}

@inproceedings{dcn2,
  title={Deformable convnets v2: More deformable, better results},
  author={Zhu, Xizhou and Hu, Han and Lin, Stephen and Dai, Jifeng},
  booktitle={Proceedings of the IEEE/CVF conference on computer vision and pattern recognition},
  pages={9308--9316},
  year={2019}
}

@inproceedings{he2020momentum,
  title={Momentum contrast for unsupervised visual representation learning},
  author={He, Kaiming and Fan, Haoqi and Wu, Yuxin and Xie, Saining and Girshick, Ross},
  booktitle={Proceedings of the IEEE/CVF conference on computer vision and pattern recognition},
  pages={9729--9738},
  year={2020}
}

@phdthesis{bias,
  title={The spatial inductive bias of deep learning},
  author={Mitchell, Benjamin R},
  year={2017},
  school={Johns Hopkins University}
}

@article{deeplearning,
  title={Deep learning},
  author={LeCun, Yann and Bengio, Yoshua and Hinton, Geoffrey},
  journal={nature},
  volume={521},
  number={7553},
  pages={436--444},
  year={2015},
  publisher={Nature Publishing Group UK London}
}

@inproceedings{krebs2017robust,
  title={Robust non-rigid registration through agent-based action learning},
  author={Krebs, Julian and Mansi, Tommaso and Delingette, Herv{\'e} and Zhang, Li and Ghesu, Florin C and Miao, Shun and Maier, Andreas K and Ayache, Nicholas and Liao, Rui and Kamen, Ali},
  booktitle={International conference on medical image computing and computer-assisted intervention},
  pages={344--352},
  year={2017},
  organization={Springer}
}

@article{HU20181,
title = {Weakly-supervised convolutional neural networks for multimodal image registration},
journal = {Medical Image Analysis},
volume = {49},
pages = {1-13},
year = {2018},
issn = {1361-8415},
doi = {https://doi.org/10.1016/j.media.2018.07.002},
url = {https://www.sciencedirect.com/science/article/pii/S1361841518301051},
author = {Yipeng Hu and Marc Modat and Eli Gibson and Wenqi Li and Nooshin Ghavami and Ester Bonmati and Guotai Wang and Steven Bandula and Caroline M. Moore and Mark Emberton and Sébastien Ourselin and J. Alison Noble and Dean C. Barratt and Tom Vercauteren}
}

@article{jaderberg2014speeding,
  title={Speeding up convolutional neural networks with low rank expansions},
  author={Jaderberg, Max and Vedaldi, Andrea and Zisserman, Andrew},
  journal={arXiv preprint arXiv:1405.3866},
  year={2014}
}

@article{yuan2006model,
  title={Model selection and estimation in regression with grouped variables},
  author={Yuan, Ming and Lin, Yi},
  journal={Journal of the Royal Statistical Society Series B: Statistical Methodology},
  volume={68},
  number={1},
  pages={49--67},
  year={2006},
  publisher={Oxford University Press}
}

@article{bengio2013representation,
  title={Representation learning: A review and new perspectives},
  author={Bengio, Yoshua and Courville, Aaron and Vincent, Pascal},
  journal={IEEE transactions on pattern analysis and machine intelligence},
  volume={35},
  number={8},
  pages={1798--1828},
  year={2013},
  publisher={IEEE}
}

@inproceedings{modet,
  title={Modet: Learning deformable image registration via motion decomposition transformer},
  author={Wang, Haiqiao and Ni, Dong and Wang, Yi},
  booktitle={International Conference on Medical Image Computing and Computer-Assisted Intervention},
  pages={740--749},
  year={2023},
  organization={Springer}
}

@article{vit,
  title={An image is worth 16x16 words: Transformers for image recognition at scale},
  author={Dosovitskiy, Alexey and Beyer, Lucas and Kolesnikov, Alexander and Weissenborn, Dirk and Zhai, Xiaohua and Unterthiner, Thomas and Dehghani, Mostafa and Minderer, Matthias and Heigold, Georg and Gelly, Sylvain and others},
  journal={arXiv preprint arXiv:2010.11929},
  year={2020}
}

@inproceedings{dcn,
  title={Deformable convolutional networks},
  author={Dai, Jifeng and Qi, Haozhi and Xiong, Yuwen and Li, Yi and Zhang, Guodong and Hu, Han and Wei, Yichen},
  booktitle={Proceedings of the IEEE international conference on computer vision},
  pages={764--773},
  year={2017}
}

@inproceedings{nnwnet,
  title={nnWNet: Rethinking the Use of Transformers in Biomedical Image Segmentation and Calling for a Unified Evaluation Benchmark},
  author={Zhou, Yanfeng and Li, Lingrui and Lu, Le and Xu, Minfeng},
  booktitle={Proceedings of the Computer Vision and Pattern Recognition Conference},
  pages={20852--20862},
  year={2025}
}

@article{cnn,
  title={Gradient-based learning applied to document recognition},
  author={LeCun, Yann and Bottou, L{\'e}on and Bengio, Yoshua and Haffner, Patrick},
  journal={Proceedings of the IEEE},
  volume={86},
  number={11},
  pages={2278--2324},
  year={2002},
  publisher={Ieee}
}

@inproceedings{swin_voxelmorph,
  title={Swin-voxelmorph: A symmetric unsupervised learning model for deformable medical image registration using swin transformer},
  author={Zhu, Yongpei and Lu, Shi},
  booktitle={International Conference on Medical Image Computing and Computer-Assisted Intervention},
  pages={78--87},
  year={2022},
  organization={Springer}
}

@article{transmorph,
  title={Transmorph: Transformer for unsupervised medical image registration},
  author={Chen, Junyu and Frey, Eric C and He, Yufan and Segars, William P and Li, Ye and Du, Yong},
  journal={Medical image analysis},
  volume={82},
  pages={102615},
  year={2022},
  publisher={Elsevier}
}

@inproceedings{glu,
  title={GLU-Net: Global-local universal network for dense flow and correspondences},
  author={Truong, Prune and Danelljan, Martin and Timofte, Radu},
  booktitle={Proceedings of the IEEE/CVF conference on computer vision and pattern recognition},
  pages={6258--6268},
  year={2020}
}

@article{vit-v-net,
  title={Vit-v-net: Vision transformer for unsupervised volumetric medical image registration},
  author={Chen, Junyu and He, Yufan and Frey, Eric C and Li, Ye and Du, Yong},
  journal={arXiv preprint arXiv:2104.06468},
  year={2021}
}

@inproceedings{xmorpher,
  title={Xmorpher: Full transformer for deformable medical image registration via cross attention},
  author={Shi, Jiacheng and He, Yuting and Kong, Youyong and Coatrieux, Jean-Louis and Shu, Huazhong and Yang, Guanyu and Li, Shuo},
  booktitle={International Conference on Medical Image Computing and Computer-Assisted Intervention},
  pages={217--226},
  year={2022},
  organization={Springer}
}

@article{attention,
  title={Attention is all you need},
  author={Vaswani, Ashish and Shazeer, Noam and Parmar, Niki and Uszkoreit, Jakob and Jones, Llion and Gomez, Aidan N and Kaiser, {\L}ukasz and Polosukhin, Illia},
  journal={Advances in neural information processing systems},
  volume={30},
  year={2017}
}
}

\clearpage
\setcounter{page}{1}
\maketitlesupplementary

\section{More serious combinatorial explosion problem in registration}

Since DMIR needs to model feature relationships between two images, the possible feature combinations are far more numerous than those in a single input task (classification, segmentation), thus leading to a more serious combinatorial explosion problem. To illustrate and rigorously prove this point, we take segmentation (a typical single input task) as an example and compare the complexity of these two tasks through mathematical methods. It is demonstrated that the combinatorial explosion in the registration task is significantly more serious than that in segmentation.

\textbf{Complexity of medical image segmentation:} Segmentation requires assigning a label to each pixel. Suppose each pixel has $L$ possible labels, then the total number of possible segmentation configurations is
\begin{equation}
|C| = L^N = L^{H W}.
\end{equation}
which is exponential growth in the number of pixels $N$.

\textbf{Complexity of deformable medical image registration:} DMIR aims to find correspondences between features in two or more images, often requiring pairwise or combination f features. Given two images, each with $N = H \times W$ feature points, the goal is to identify correct matches. For each feature point $f_i$ in the source image, there exists a candidate set of potential matches in the target image. Let the average size of this candidate set be proportional to the image size, modeled as $c = \alpha N$, where $\alpha$ is the average number of possible feature relationships for each pair of feature combinations. Assuming independent choices for each feature, the total number of possible feature combinations is:
\begin{equation}
|\mathcal{H}| = c^N = (\alpha H W )^{H W}.
\end{equation}
This expresses an exponential growth of the possible feature combinations with respect to $H \times W$.

\textbf{Proof of the more serious combinatorial explosion problems in registration:} The total number of possible segmentation configurations is $|C| = L^N = L^{H W}$, where $L$ is the fixed number of labels, and $N = H \times W$ is the number of pixels. The total number of possible feature combinations in registration is $|\mathcal{H}| = c^N = (\alpha N - 1)^N = (\alpha H W - 1)^{H W}$, where $c = \alpha N - 1$ is the average candidate set size per feature point and grows linearly with $N$. To compare their growth rates, consider the logarithms of these quantities:
\begin{equation}
\log |C| = N \log L,
\end{equation}
and
\begin{equation}
\log |\mathcal{H}| = N \log (\alpha N - 1).
\end{equation}
The ratio of these logarithms is
\begin{equation}
R = \frac{\log |\mathcal{H}|}{\log |C|} = \frac{N \log (\alpha N - 1)}{N \log L} = \frac{\log (\alpha N - 1)}{\log L}.
\end{equation}
As $N$ grows large,
\begin{equation}
\log (\alpha N - 1) \approx \log N + \log \alpha \to \infty,
\end{equation}
while $\log L$ is constant. Therefore,
\begin{equation}
\lim_{N \to \infty} R = \lim_{N \to \infty} \frac{\log (\alpha N - 1)}{\log L} = \infty.
\end{equation}
This shows that the logarithm of the registration complexity grows faster than that of segmentation complexity, implying
\begin{equation}
|\mathcal{H}| \gg |C| \quad \text{as} \quad N \to \infty.
\end{equation}
In other words, the combinatorial explosion in registration tasks is more serious than in segmentation tasks.

\section{Technical Details}

\subsection{Details of registration architecture}

\begin{figure*}[t] 
  \centering
  \includegraphics[width=\textwidth]{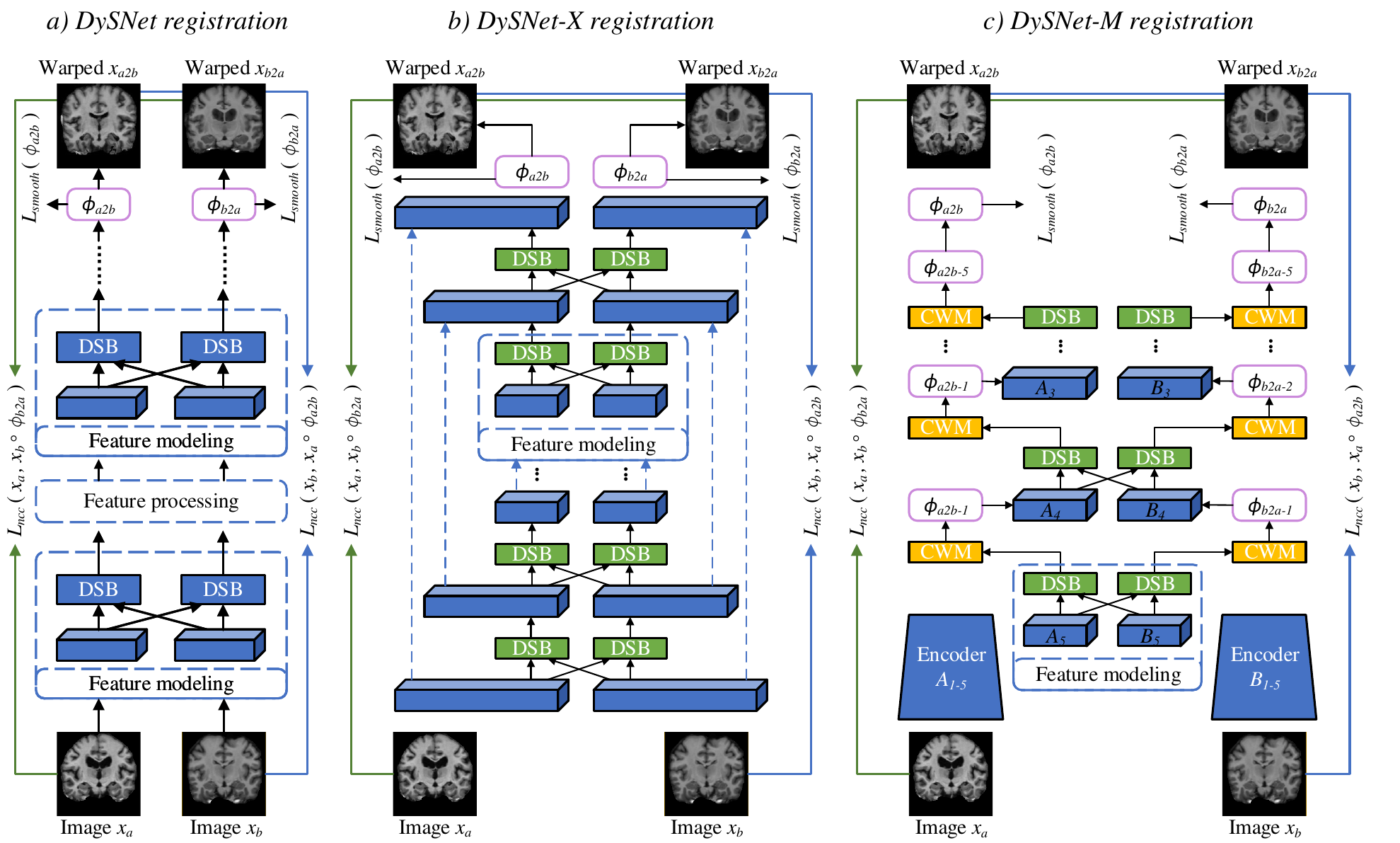}
  \caption{The overall architecture of DySNet and its two instantiated networks are presented. a) DySNet models the input features through alternately stacked DSB modules. Based on this general network, two instantiated networks are derived: b) DySNet-X (based on Xmorpher), c) DySNet-M (based on ModeT).}
  \label{f1}
\end{figure*}

\textbf{Details of DySNet:} As show in Fig.\ref{f1}-a, DySNet is designed as a versatile and flexible bidirectional registration architecture. It first forms a bidirectional feature modeling component through alternating DSB modules, and then stacks this feature modeling component to obtain the overall network. Components can use different feature processing methods among themselves. This module alternately processes the features from the two input images to model complex feature relationships. Following Xmorpher~\cite{xmorpher}, $\mathcal{L}_{reg}$ is a deformable registration loss composed of smoothness loss $\mathcal{L}_{smo}$ (Jacobian matrix), which preserves deformation topology, and similarity loss $\mathcal{L}_{sim}$ (mean dice similarity coefficients ), which aligns similar regions.

\textbf{Details of DySNet-X:} As show in Fig.\ref{f1}-b, DySNet-X is a variant of DySNet. In the original Xmorpher~\cite{xmorpher} architecture, the CAT block is replaced by the proposed DSB for feature modeling. By replacing the CAT~\cite{xmorpher} block with DSB, DySNet-X inherits the bidirectional and symmetric modeling advantages of DySNet and performs registration based on the dynamic feature modeling provided by DSB. This leads to improved registration accuracy and more consistent deformation fields, as DSB significantly strengthens the symmetry prior and can more effectively perform feature modeling.

\textbf{Details of DySNet-M:} As show in Fig.\ref{f1}-c, DySNet-M is achieved by a symmetric pyramid registration structure (ModeT~\cite{modet}). It uses an encoder with 5 convolutional layers to extract hierarchical features, producing feature maps $A_1\!-\!A_5$ and $B_1\!-\!B_5$. Feature map $A_5$ and $B_5$ go through DSB and CWM~\cite{modet} to obtain deformation $\phi_{a2b-1}$ of $A_4$. The deformed $A_4$ and $B_4$ input DSB and CWM to generate subfields $\phi_{a2b-2}$. Similar steps apply to $B_3$ and $A_3$. Finally, the $\phi_{a2b}$ obtained from the network warps the image $x_a$ to obtain the registration result, and the processing of image $x_b$ is carried out in the same way.

\begin{figure*}[t] 
  \centering
  \includegraphics[width=\textwidth]{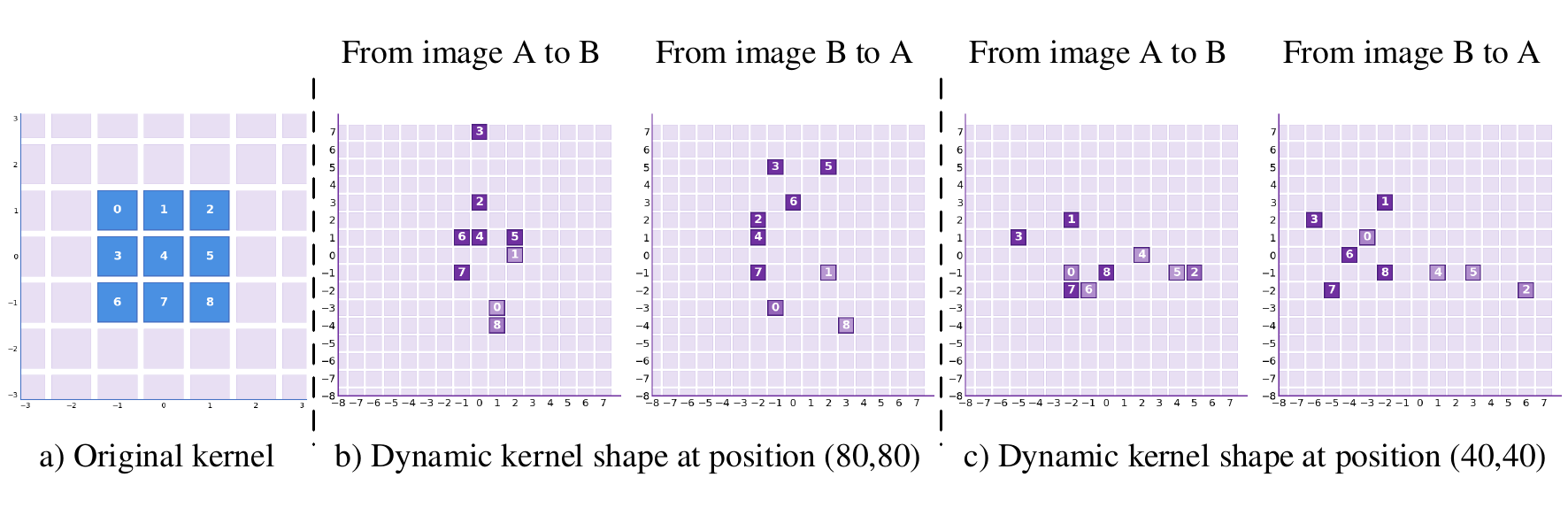}
  \caption{The figure shows the deformation of the convolution kernel. a) Illustrates the original fixed 3x3 convolution kernel. b) and c) are the visualizations of the dynamic kernels at the first layer position (80, 80) and the second layer position (40, 40) of the DySNet-X network, respectively. The shape of the kernel is adaptively adjusted according to the input features.}
  \label{f2}
\end{figure*}

\subsection{Details of DySNet's attention mechanisms}

Building upon the formulations in Sections 3.1 and 3.2, we further elaborate on the details and properties of the dynamic deformable attention mechanisms in DySNet.

\textbf{Multi-head attention projection and reshaping:} Given input features $f^a, f^b \in \mathbb{R}^{B \times C \times H \times W}$, the linear projections for queries, keys, and values are implemented as convolutional layers followed by reshaping:

\begin{equation}
Q = \mathrm{Conv}_q(f^a), \quad K = \mathrm{Conv}_k(f^b), \quad V = \mathrm{Conv}_v(f^b),
\end{equation}
where each of $Q, K, V \in \mathbb{R}^{B \times C \times H \times W}$ is reshaped to multi-head representation:

\begin{equation}
Q, K, V \to \mathbb{R}^{B \times d \times h \times H \times W},
\end{equation}
with $d = \frac{C}{h}$ the per-head channel dimension and $h$ the number of attention heads.

\textbf{Interpolation as continuous sampling operator:} The interpolation function $I(\cdot)$ used to sample deformed keys and values at coordinates $D_{N^{d}}(i)_j \in \mathbb{R}^2$ is implemented as bilinear interpolation, which can be expressed as:

\begin{equation}
I(F, p) = \sum_{q \in \mathcal{N}(p)} w_q(p) F(q),
\end{equation}
where $\mathcal{N}(p)$ are the four integer grid neighbors of the fractional coordinate $p$, and $w_q(p)$ are bilinear interpolation weights satisfying $\sum_q w_q(p) = 1$. This ensures the interpolation is continuous and differentiable with respect to $p$, enabling backpropagation through the dynamic offset $\Delta i$.

\textbf{Offset prediction network and smoothness:} The offset prediction network $\theta_{\mathrm{offset}}$ is a small CNN module taking concatenated features $X = [f^a, f^b] \in \mathbb{R}^{B \times 2C \times H \times W}$ and predicting offsets:

\begin{equation}
\Delta i = \theta_{\mathrm{offset}}(X) \in \mathbb{R}^{B \times |U_{N^d}| \times d \times H \times W}.
\end{equation}
The smoothness of $\Delta i$ over spatial positions is implicitly encouraged by the convolutional nature of $\theta_{\mathrm{offset}}$ and regularization losses (if any), leading to locally coherent deformations.

\textbf{Joint optimization perspective:} The parameters of $\theta_{\mathrm{offset}}$, $\mathrm{Conv}_q$, $\mathrm{Conv}_k$, $\mathrm{Conv}_v$, and output projection are optimized end-to-end, enabling the network to jointly learn optimal dynamic receptive fields $D_{N^{d}}(i)$ and corresponding attention weights $\rho_{N^{d},1}^i$ that minimize the target loss. This synergy allows DySNet to adaptively focus on spatially relevant features and handle local deformations with precise and smooth spatial attention.


\section{Experiment Details}

\subsection{Datasets}

\textbf{MM-WHS~\cite{MM-WHS}} The dataset consists of 120 three-dimensional cardiac image volumes, covering 60 CT scans and 60 MRI scans, all of which were collected from real clinical environments. The images in the dataset are all labeled with seven major anatomical structures of the heart, including the left and right ventricles, left and right atria, left ventricular myocardium, ascending aorta, and pulmonary artery. In this study, the CT modality was used as the training set, and the cardiac region in the images was cropped and resampled to a size of 144×144×128 before being normalized.

\textbf{ASOCA~\cite{ASOCA}} The data focus on the automatic segmentation of coronary arteries, including 60 cases of CT coronary angiography (CTCA), among which 30 cases are normal without any lesions, and 30 cases are patients with different degrees of coronary artery diseases. The dataset is equipped with a complete coronary artery tree structure annotated jointly by multiple experts, covering the entire tree structure of the left and right coronary arteries. The cardiac region in the images was cropped and resampled to a size of 144×144×128 before being normalized.

\textbf{CAT08~\cite{CAT08}} The dataset consists of 32 three-dimensional CT images with labels indicating the structure of the heart, covering the annotations of the main chambers of the heart and major blood vessels. The cardiac region in the images was cropped and resampled to a size of 144×144×128 before being normalized.

\textbf{PPMI~\cite{ppmi}} It is extracted from the PPMI database which is a large Parkinson progression marker initiative database, for 837 T1 brain MR volumes. We resize and crop 160×160×128 volumes on the brain regions, and then normalize the intensity. We also extract the brain regions via HD-BET~\cite{HD-BET} to avoid the interference of background.

\textbf{CANDI~\cite{candi}} The Child and Adolescent Neuro Development Initiative (CANDI) dataset has 103 T1 brain MR volumes from 57 males and 46 females. Totally 28 brain tissue regions are annotated for masks. We resize and crop 160×160×128 volumes on the brain regions, and then normalize the intensity.

\textbf{OASIS-1} This dataset comprises a cross-sectional sample of 416 participants aged between 18 and 96. Each participant has 3 to 4 T1-weighted MRI scans obtained from individual single scans. The brain region was cropped and resampled to a size of 160×160, and the data has already been standardized by the official.

\subsection{Implementations}

Our DySNet were conducted on NVIDIA RTX 6000 with 24\,GB of memory, using PyTorch~\cite{pytorch}. We used AdamW \cite{adamw} as the optimizer with an initial learning rate of $10^{-4}$. The total number of training rounds is 200 with a batch size of 1. To conduct a fair comparison, all the methods employed in our experiment either adopted the same hyperparameter settings or adhered to the experimental settings provided in the original papers. 
\section{More Framework Analysis and Results}

\begin{figure}[t] 
  \centering
  \includegraphics[width=1\linewidth]{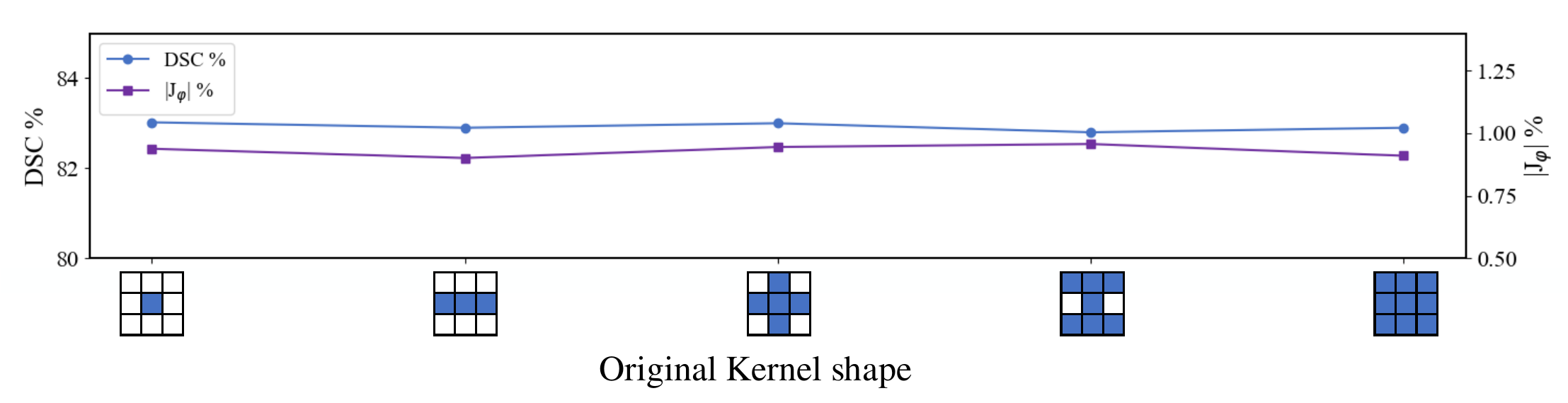}
  \caption{The figure shows the performance of the DySNet network under different shapes of the original kernels and the number of points. The x-axis represents the shape of the original kernel, and the y-axis represents the DSC \% and $|J_{\phi}|\% $. The experimental results demonstrate that, despite the changes in the shape and number of points of the original kernel, DySNet still maintain stable and excellent performance, verifying its ability to dynamically adjust the receptive field and weights.}
  \label{f3}
\end{figure}

\subsection{Visualization of dynamic kernel}

As shown in the Fig.\ref{f2}, the dynamic kernel of DySNet-X adaptively adjust the shape and weights (shade of purple indicates the magnitude of the weight) of the window according to the different input data, achieving more flexible feature modeling. The shape of the dynamic kernel at the first layer of DySNet-X at the position (80, 80) significantly changed with the exchange of the registered images, demonstrating its ability to accurately capture local features. The dynamic kernel at the position (40, 40) of DySNet-X in the second layer showed different deformations, indicating that the network performed adaptive modeling of the structural information of the image at a deeper level. This dynamic adjustment mechanism helps to enhance the model's adaptability to spatial deformation and complex textures, thereby strengthening the overall feature modeling ability.

\begin{figure}[t] 
  \centering
  \includegraphics[width=1\linewidth]{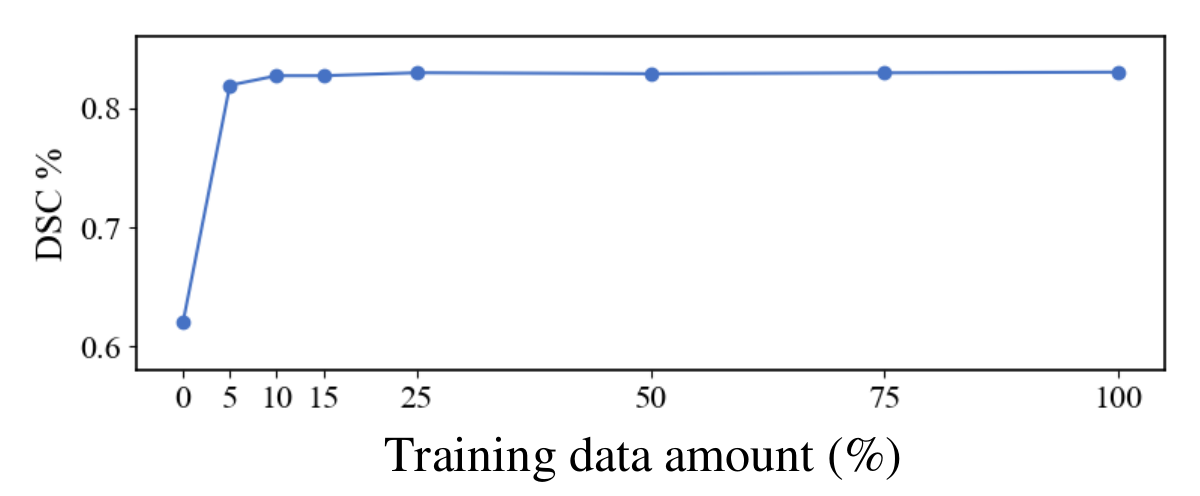}
  \caption{The figure shows the performance changes of DySNet in the 2D brain assessment when using different amounts of training data (\%).}
  \label{f4}
\end{figure}

\subsection{Analysis of original kernel shape}

As shown in the Fig.\ref{f3}, in addition to adjusting the size of the kernel, we also verified the dynamic adaptability of the DySNet network when dealing with different original kernel shapes through this experiment. The results indicated that the performance of DySNet changed very little under different kernel shapes and the number of points, fully demonstrating the dynamic nature and robustness of the network. This not only reduces the reliance on the selection of kernel shapes but also enhances the generalization ability of the model, reflecting the advantage of dynamic in the design concept of DySNet.

\subsection{Analysis of training data amount}
As shown in Fig.\ref{f4}, we evaluate the variation of our DySNet’s performance with the enlarging of the training data amount on our 2D brain evaluation. Our DySNet brings a significant improvement even though only 5\% training data is involved. When further enlarging the training dataset, the gain of performance gradually decreases owing to the similarity of medical images in the training dataset. Fortunately, our DySNet possesses a powerful ability to model dynamic features and can effectively learn from details. As a result, the performance of this model is still improving gradually.

\end{document}